\documentclass[runningheads]{llncs}

\usepackage{graphicx}
\usepackage{amsmath}
\usepackage{amssymb}
\usepackage{siunitx}
\usepackage{hyperref}
\usepackage[nameinlink, capitalize, noabbrev]{cleveref}
\usepackage{tabularray}
\usepackage{diagbox}
\usepackage{subcaption}
\usepackage{orcidlink}

\DeclareSIUnit\ppm{ppm}
\DeclareSIUnit\Molar{M}

\newcommand{\Dw}{\Delta\omega}
\newcommand{\dw}{\delta\omega_i}
\newcommand{\w}{\omega_1}
\newcommand{\mtr}{MTR(\Dw)}
\newcommand{\mtrasym}{MTR_{asym}(\Dw)}
\newcommand{\mtrrex}{MTR_{Rex}(\Dw)}
\newcommand{\rex}{R_{ex}^{(i)}(\Dw)}

\begin{document}

\title{Transformer-based Parameter Fitting of Models derived from Bloch-McConnell Equations for CEST MRI Analysis}
\titlerunning{Transformer-based CEST MRI Analysis}

\author{Christof Duhme\inst{1}\orcidlink{0009-0004-1853-2862} \and
Chris Lippe\inst{1}\orcidlink{0009-0004-3006-3915} \and
Verena Hoerr\inst{1}\orcidlink{0000-0002-5301-3636} \and
Xiaoyi Jiang\inst{1}\orcidlink{0000-0001-7678-9528}}
\authorrunning{C. Duhme et al.}
\institute{University of Münster, Germany}

\maketitle 

\begin{abstract}
Chemical exchange saturation transfer (CEST) MRI is a non-invasive imaging modality for detecting metabolites.
It offers higher resolution and sensitivity compared to conventional magnetic resonance spectroscopy (MRS).
However, quantification of CEST data is challenging because the measured signal results from a complex interplay of many physiological variables.
Here, we introduce a transformer-based neural network to fit parameters such as metabolite concentrations, exchange and relaxation rates of a physical model derived from Bloch-McConnell equations to in-vitro CEST spectra.
We show that our self-supervised trained neural network clearly outperforms the solution of classical gradient-based solver.

\end{abstract}

\textbf{\textcopyright 2025 Springer.} The Version of Record of this contribution is published in Machine Learning in Medical Imaging. MLMI 2024. Lecture Notes in Computer Science, vol 15242., and is available online at \url{https://doi.org/10.1007/978-3-031-73290-4_11}.

\section{Introduction}
Since Ward et al. \cite{ward2000cest} proposed the use of Chemical Exchange dependent Saturation Transfer (CEST) based contrast agents, CEST MRI has emerged as a promising technique for non-invasive metabolite detection \cite{vanZijl2011CEST}.
The principle of CEST involves using radiofrequency (RF) pulses to saturate specific molecules capable of exchanging protons with water, indirectly measuring their concentration by the resultant decrease in water signal in MRI.
From a data science perspective, this can be understood as an additional dimension of spectroscopic information complementing the imaging data.
This technique significantly enhances the detectability of these molecules, present in very low concentrations, by continuously transferring saturation to water, amplifying signal changes by a factor of up to $100$ compared to conventional methods \cite{vanZijl2011CEST}.
For CEST analysis, two main approaches are employed: model-free and model-based methods \cite{foo2020analysis}. Model-free methods focus on deriving image contrasts at specific frequency offsets or ranges. Model-based analysis, on the other hand, involves fitting parameters of models, such as numerical solutions to Bloch-McConnell equations, to measured spectra, aiding in the extraction of physiological parameters and enhancing imaging contrast by isolating individual contributions \cite{foo2020analysis}.

Model-based CEST data analysis generally faces two types of challenges: overly complex physical models, leading to inconsistent fitting results, and instability of solvers. In this work, we will address the latter by introducing a self-supervised transformer-based neural network to fit parameters of a Lorentzian model (\cref{eq:lorentzian-model}) and, to the best of our knowledge, for the first time two physical models derived from Bloch-McConnell equations namely an analytical $Z$ (\cref{eq:z-model}) and $MTR_{Rex}$ model (\cref{eq:mtrrex-model}).
We show that our network performs better and more consistent than classical solvers.

\section{Related Work}\label{sec:related_work}
A variety of model-based methods have been developed to analyze CEST data.
Foo et al. \cite{foo2020analysis} discussed the applicability of multi-Lorentzian fitting in CEST MRI analysis across different medical applications, highlighting the versatility of model-based approaches. Zaiss et al.
\cite{zaiss2011quantitative} presented a method for $Z$-spectra analysis that quantitatively separates CEST effects from magnetization transfer and spillover effects, providing a foundation for precise CEST effect quantification.
In the realm of specific applications, Zhou et al. \cite{zhou2017quantitative} proposed the IDEAL fitting method based on initial values from multi-pool Lorentzian fitting for glioma quantification in CEST MRI.
Moreover, Zhang et al. \cite{zhang2023mri} developed a quantitative method using a 5-pool Lorentzian fitting to detect tissue acidosis following an acute stroke, showcasing the potential of CEST MRI in acute medical conditions.
Lastly, Zaiss et al. \cite{zaiss2019quantification} optimized glucoCEST MRI experiments at various field strengths by employing the Bloch-McConnell equations to account for hydroxyl exchange of D-Glucose under physiological conditions, indicating the precision and adaptability of CEST MRI for metabolic studies.

Lately, deep learning is more and more applied to the field of CEST MRI.
Glang et al. \cite{glang2020deepcest} predicted the parameters for a Lorentzian model with a deep learning network and provided uncertainty quantification.
Huang et al. \cite{huang2022deep} applied this model to imaging of mice brains with Alzheimer's disease.
Singh et al. \cite{singh2023bloch} used recurrent neural networks for Bloch simulation and MR fingerprinting reconstruction.
To reconstruct undersampled multi-coil CEST data, Xu et al. \cite{xu2024accelerating} used a variational neural network. They also synthesized multi-coil CEST data from anatomical MRI data.
Xiao et al. \cite{xiao2024deep} used a recurrent neural network for upsampling of sparse $Z$-spectra.
Cohen et al. \cite{cohen2023global} used deep learning to optimize the magnetic resonance fingerprinting schedule for CEST MRI to reduce long scan times and simultaneously yield better results.

\section{Physical Background of CEST MRI}\label{sec:physical_background}
A typical CEST experiment works as follows: one first conducts an unsaturated reference image showing localized signal intensities $S_0$. Then, one acquires a series of images with radio frequency (RF) saturation pulses applied prior to RF excitation. The frequency offset $\Dw$ of the saturation pulse frequency to the water resonance frequency is varied for each image of the sequence. Measured signal intensities $S(\Dw)$ are then normalized by $S_0$ resulting in a functional relationship $Z(\Dw) = S(\Dw)/S_0\in (0,1)$, also referred to as $Z$-spectrum.

The $Z$-spectrum depends on experimental and physiological parameters. The relevant experimental parameters are saturation type (continuous wave or pulsed), time ($t_{sat}$) and strength/amplitude ($B_1$ or $\w=\gamma B_1$, where $\gamma$ is the gyromagnetic ration of $^1$H). In the following, we only consider continuous wave saturation and long saturation times, sufficient to reach steady-state magnetization. This substantially simplifies the underlying theory.

The physiological parameters of interest are the longitudinal and transversal relaxation rates of the water pool $R_{1a}$ and $R_{2a}$, the proton fractions $f_i$, exchange rates $k_i$, transversal relaxation rates $R_{2i}$ and the resonance frequency offsets $\dw$ of the contributing solute pools $i = 1\dots n$. For a detailed discussion of all parameters that influence $Z$-spectra, see \cite{zaiss2013chemical}.

As introduced, model-free and model-based analysis methods for CEST data exist. \cite{foo2020analysis}.
The following subsections give an overview of available metrics and models to quantify CEST data.

\subsection{CEST metrics}
The magnetization transfer ratio $MTR$ is calculated as:
\begin{equation}\label{eq:mtr}
    \mtr = 1 - Z(\Dw)
\end{equation}
The asymmetric magnetization transfer ratio $MTR_{asym}$ suppresses symmetric parts of the $Z$-spectra which mainly arise from unwanted direct water saturation (DS) and magnetization transfer (MT) contributions:
\begin{equation}\label{eq:mtrasym}
    \mtrasym = Z(-\Dw) - Z(\Dw)
\end{equation}
As $Z$ is a reciprocal quantity, $MTR_{asym}$ does not suppress all symmetric contributions. Therefore, $MTR_{Rex}$ has been introduced which is given by:
\begin{equation}\label{eq:mtrrex}
    \mtrrex = \frac{1}{Z(\Dw)} - \frac{1}{Z(-\Dw)}
\end{equation}

\subsection{CEST models}
The most general model we consider is the approximate steady-state solution for the $Z$-spectrum from Zaiss and Bachert \cite{zaiss2013chemical} derived from BM equations. It holds under the assumption that $t_{sat}>>T_2$, $\w>>1/T_2$, where $T_2$ is the observed transversal relaxation time, asymmetric populations (i.\,e. $f_i<<1$), negligible longitudinal relaxation rates of the pools $R_{1i}<<k_i$ and reads:
\begin{equation}\label{eq:z-model}
    Z(\Dw) = \frac{R_{1a} \Dw^2}{R_{1a} \Dw^2 + R_{2a} \cdot \w^2 + (\w^2 + \Delta w^2) \sum_i \rex},
\end{equation}
where
\begin{equation}
    \rex = \frac{f_i \cdot \w^2}{\Gamma_i^2/4 + (\Dw - \dw)^2} \biggl( R_{2i} + k_i \frac{\dw^2 + R_{2i} (R_{2i} + k_i)}{\w^2 + \Dw^2} \biggr)
\end{equation}
and further
\begin{equation}
    \Gamma_i^2/4 = \frac{R_{2i} + k_i}{k_i} \w^2 + (R_{2i} + k_i)^2.
\end{equation}
It should be noted that only the ratios $\frac{R_{2a}}{R_{1a}}$ and $\frac{\rex}{R_{1a}}$ are unique and can be fitted. For constant $R_{1a}$, the ratios $\frac{f_i}{R_{1a}}$ are directly proportional to the metabolite concentrations. Using the same model for $\rex$, $\mtrrex$ as in \cref{eq:mtrrex} can also be fitted:
\begin{equation}\label{eq:mtrrex-model}
    MTR_{Rex}(\Dw) \cdot\frac{\Dw^2}{\Dw^2 + \w^2}= \frac{\sum_i \rex}{R_{1a}},
\end{equation}
where the left hand side is calculated numerically and the right hand side is fitted.
Finally, we consider a more simple Lorentzian multi-pool model:
\begin{equation}\label{eq:lorentzian-model}
    \mtr = \sum_i\frac{a_i\cdot\Gamma_i^2/4}{(\Dw - \dw)^2 + \Gamma_i^2/4},
\end{equation}
where $a_i$ and $\Gamma_i$ are the amplitudes and widths of the respective pools.
This model is effectively a further approximation of \cref{eq:z-model} under the assumption of large $\dw$ and small $\rex$. These assumptions typically do not hold for many solutes of interest, such as lactate, glucose, amide and amine, but this model has still proven to be useful in improving separation of individual contributions \cite{zhang2023mri,zhou2017quantitative}. CEST contrast is given by the amplitudes or area under the curves (AUC), which are expected to be proportional to the proton fractions $f_i$.

\section{Fitting the Physical Models}\label{sec:methods}
From \cref{sec:physical_background} we now use the Lorentzian model (\cref{eq:lorentzian-model}), the analytical $Z$ model (\cref{eq:z-model}) and the $MTR_{R_{ex}}$ model (\cref{eq:mtrrex-model}).
Usually, solutions to the proposed models are found using iterative solvers.
As we know reasonable bounds for our parameters, we use a L-BFGS-B \cite{byrd1995limited,zhu1997algorithm}, Nelder-Mead \cite{nelder1965simplex} and Powell \cite{powell1964efficient} solver with squared error as the minimizing function.

\subsection{Transformer-based Neural Network}
Our neural network is an encoder-decoder architecture.
The encoder is a transformer with 8 layers and multi-head attention \cite{vaswani2017attention} with 8 attention heads, hidden size 1024 and MLP dimension 1024.
The decoder is a convolutional network with $3 \times 3$ convolution with 512, 256, 128 and 64 channels with a MLP head that predicts the parameters of the used physical model in the specified bounds.

We chose a transformer-based architecture as it can model the long-range causal relationship between different frequencies of the CEST spectra via the attention mechanism.
Preliminary experiments during development with convolutional encoders validated this assumptions as those networks failed to learn meaningful information.
The size of our model was determined by starting with a small model and scaling it up until it was capable of learning the structure of CEST data.
Thus, we prevented using an oversized model which would have been prone to overfitting.

Let $f(x)$ be the output of our network for the input CEST curves $x$. Let $M$ be the physical model with parameter bounds $[c_M - d_M, c_M + d_M]$ given by $c_M$ the vector of centers and $d_M$ the vector of deviations. We calculate the vector of parameters $p_M(x)$ for $M$ with
\begin{equation}
    p_M(x) = c_M + d_M \cdot \tanh f(x).
\end{equation}
The reconstruction loss $L_g(x)$ for some loss function $g$ is then given by
\begin{equation}
    L_g(x) = g(M(p_M(x)), x).
\end{equation}
Thus, we incorporate the constraints of the physical model into our network via the loss function.
We choose $g$ as the mean squared error.

We do not use the concentrations of the solutes as ground truth data to regularize our model as these would not be available when expanding the models to \textit{in vivo} data.
We only want to learn a good fit by using the intrinsic characteristics of our data.

\subsection{Performance analysis}
Evaluation is done comparing the ground truth concentrations and physical parameters or just model parameters in the case of the Lorentzian model that scale linear with regard to the proton fractions.
For the Lorentzian model, we examine the amplitude and area under the curves.
For the models based on the Bloch-McConnell equations, we look at the fractions $\frac{f_i}{R_1a}$.
We then fit a linear regression model to the ground truth concentrations and the predicted values using Ordinary Least Squares while forcing the intercept to 0 and then calculate the $R^2$ score between our obtained parameters and the linear regression.

Let $y \in \mathbb{R}^n$ be the true values (extracted model parameters) and $\hat{y} \in \mathbb{R}^n$ the predicted values (regression result).
The $R^2$ score for intercept 0 is defined as:
\begin{equation}\label{eq:r2_intercept}
    R^2(y, \hat{y}) = 1 - \frac{\sum (y - \hat{y})^2}{\sum y^2}
\end{equation}
It measures the goodness of fit and gives us a percentage of how much of the variability in the predicted parameters is accounted for by the linear regression model.

\section{Experiments \& Results}\label{sec:results}
For the sake of reproducibility, all source code is available at \url{https://zivgitlab.uni-muenster.de/ag-pria/cest}.

\subsection{Dataset}
Our experiments were performed on a \SI{9.4}{\tesla} Bruker Biospec MR system equipped with a \SI{72}{\mm} quadrature coil. We conducted CEST RARE images at \num{129} offsets (\SIrange{-5}{5}{\ppm}) with a \SI{4}{\second} saturation pulse and 
$B_1 = 1.2\,\si{\micro\tesla}$, $1.6\,\si{\micro\tesla}$, $2.0\,\si{\micro\tesla}$ and $2.4\,\si{\micro\tesla}$
on nine mixed glucose/lactate phantoms, dissolved in sterile water at concentrations of \SI{5}{\milli\Molar}, \SI{15}{\milli\Molar} and \SI{30}{\milli\Molar}, with a pH range of \num{6.5} to \num{7} at a temperature of \SI{20}{\degreeCelsius}. We extracted 529 pixel-wise spectra from the homogeneous phantoms. Spectra were normalized using the measured signal at $\Dw=$\SI{5}{\ppm} and $B_0$-corrected using cubic spline interpolation to estimate the water frequency offset.
Compared to a synthetic dataset, our phantom dataset provides a more realistic assessment of our models feasibility as it includes measurement imperfections arising from acquisition hardware limitations and variations of experimental parameters like e.g. temperature, pH and ionic strength.

\subsection{Experimental Setup}
Our neural network is trained self-supervised with 5-fold cross-validation for 200 epochs for the Lorentzian model and 1000 for the other models with mean-squared error loss and Adam optimizer with learning rate 1e-5.
Different $B_1$ values are not considered by the Lorentzian model. Therefore, it was only trained and evaluated on a single $B_1$ value of $\SI{1.2}{\micro\tesla}$. The analytical $Z$ model and $MTR_{Rex}$ use all $B_1$ values as input and ground truth data. The solvers use the center of the given bounds as the initial guess and are run until their convergence condition is met. Initializing with random guesses did not change the results.
All experiments were performed on a workstation with Intel i7-6700K CPU, 32 GB RAM and NVIDIA RTX A4000 GPU.

\subsection{Results \& Discussion}
\cref{tab:results} shows our results.
As the L-BFGS-B solver clearly outperforms the Nelder-Mead and Powell solver, we will focus on comparing it's performance to our network's.
L-BFGS-B achieves higher $R^2$ values for glucose than for lactate for all models.
Our neural network performs the same for the $MTR_{R_{ex}}$ model but reaches higher values for lactate in the Lorentzian and analytical $Z$ model.
We can see that our network outperforms the classical iterative solver for every physical model.

The relatively high $R^2$ values for the Lorentzian model with L-BFGS-B resulted from adjusted bounds for $\Gamma^2_i$.
With the standard bound of $[0,1]$ the solver almost never converges and collapses the parameters to $0$.
With an adjusted bound of $[0.3, 0.6]$ the solver still almost always chooses the lower bound of $0.3$, but is now able to find suitable amplitudes $\alpha_i$.
Our network performs the same with both set of bounds.

\begin{table}[t]
    \caption{$R^2$ values of predicted model parameters and a linear regression model with respect to ground truth concentrations. Mean and standard deviation given.}\label{tab:results}
    \vspace{2mm}
    \centering
    \begin{tblr}{
        colspec={|l|r|r|r|r|r|r|},
        colsep = 7pt,
        hline{1,2,3,5,7,9,11} = {solid},
        vlines,
        row{3,5,7,9} = {belowsep=0pt},
        row{4,6,8,10} = {rowsep=0pt, font=\scriptsize},
    }
        \SetCell[r=2]{c} & \SetCell[c=2]{c} Lorentzian & & \SetCell[c=2]{c} analytical $Z$ & & \SetCell[c=2]{c} $MTR_{Rex}$ &\\
         & \SetCell[]{c} glucose & \SetCell[]{c} lactate & \SetCell[]{c} glucose & \SetCell[]{c} lactate & \SetCell[]{c} glucose & \SetCell[]{c} lactate\\
        \SetCell[r=2]{m} Nelder-Mead & 0.9518 & -0.3152 & 0.8348 & 0.7144 & 0.9685 & 0.8355\\
         & $\pm$ 0.0109 & $\pm$ 0.5688 & $\pm$ 0.0453 & $\pm$ 0.0349 & $\pm$ 0.0149 & $\pm$ 0.0486\\
         \SetCell[r=2]{m} Powell & 0.9049 & 0.4312 & 0.7234 & 0.8439 & 0.9944 & 0.7519\\
         & $\pm$ 0.0395 & $\pm$ 0.2157 & $\pm$ 0.0308 & $\pm$ 0.0537 & $\pm$ 0.0039 & $\pm$ 0.1728\\
         \SetCell[r=2]{m} L-BFGS-B & 0.9939 & 0.9474 & 0.9215 & 0.7237 & 0.9438 & 0.8993\\
         & $\pm$ 0.0025 & $\pm$ 0.0654 & $\pm$ 0.0216 & $\pm$ 0.0011 & $\pm$ 0.0548 & $\pm$ 0.0193\\
         \SetCell[r=2]{m} Ours & 0.9589 & 0.9967 & 0.9725 & 0.9824 & 0.9972 & 0.9722\\
         & $\pm$ 0.0120 & $\pm$ 0.0001 & $\pm$ 0.0054 & $\pm$ 0.0087 & $\pm$ 0.0019 & $\pm$ 0.0152\\
    \end{tblr}
\end{table}

\begin{figure}[t]
    \centering
    \begin{subfigure}{0.5\textwidth}
        \centering
        \includegraphics[width=0.9\textwidth]{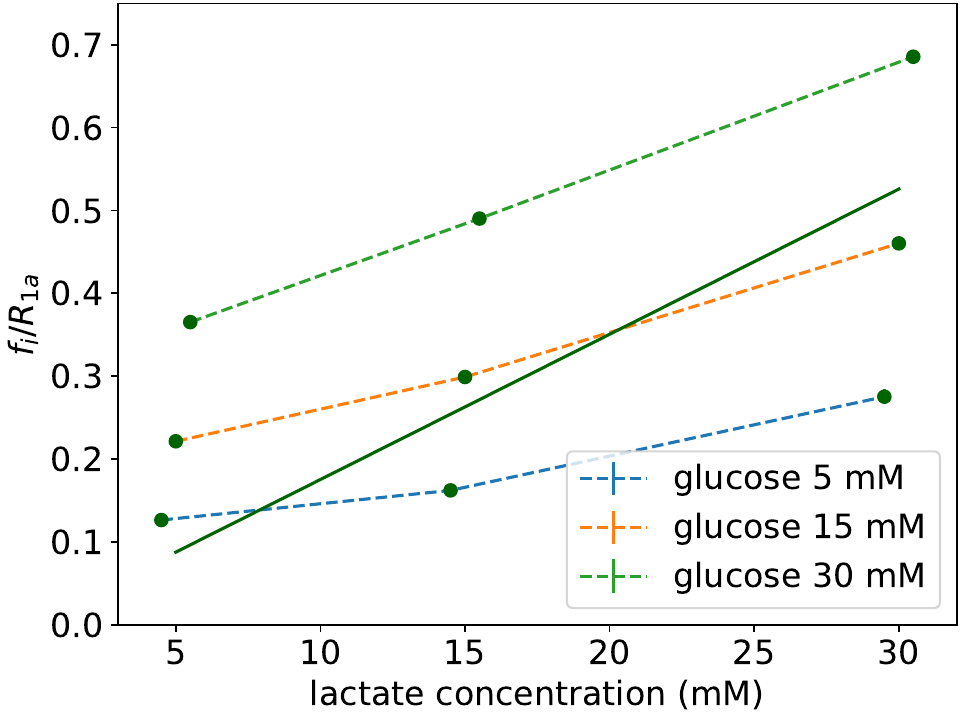}
        \caption{Our network.}
        \label{fig:example-nn}
    \end{subfigure}%
    \begin{subfigure}{0.5\textwidth}
        \centering
        \includegraphics[width=0.9\textwidth]{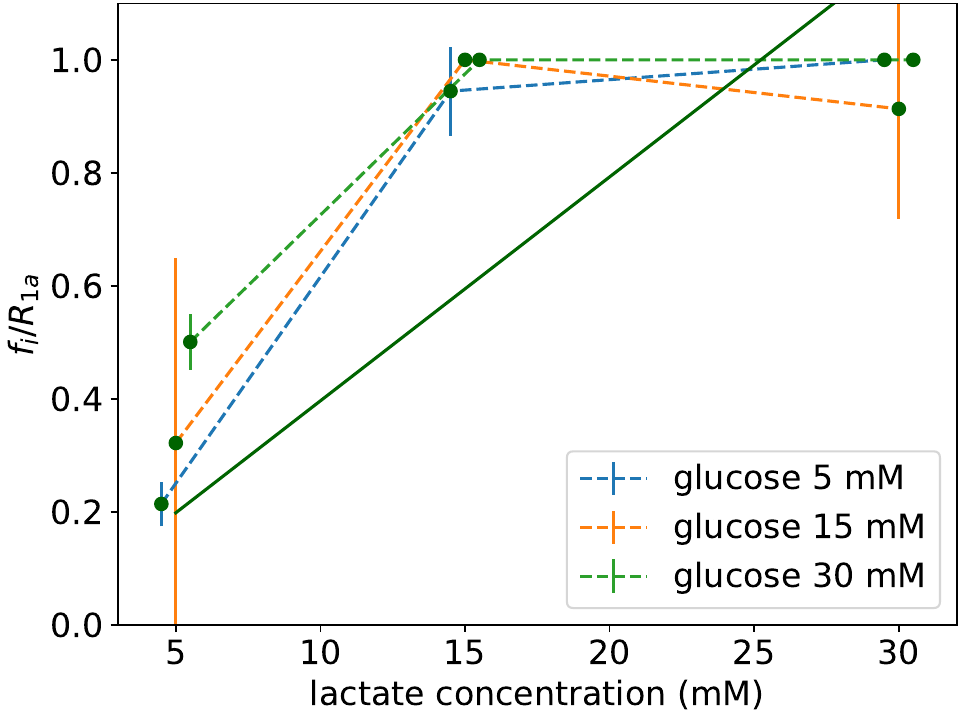}
        \caption{L-BFGS-B.}
        \label{fig:example-solver}
    \end{subfigure}
    \caption{Example plots of proton fractions for lactate for the $MTR_{R_{ex}}$ model. Dashed lines are the parameters of the model with the colors indicating the concentration of glucose, bold lines are the result of the linear regression model, errorbars are computed over the test set.}
    \label{fig:example}
\end{figure}

In the example in \cref{fig:example} we can see the neural network producing more uniform results than L-BFGS-B.
Its plots are monotonically increasing and behaving consistent over the different concentration combinations of glucose and lactate.
The solver does not behave linearly for lactate in this example (see \cref{fig:example-solver}).
Overall, both methods sort the different concentration combinations in the same order with respect to the scale of concentration of the other solute, respectively.
Meaning, the parameters increase for a given concentration if the concentration of the other solute rises.

The neural network performs very well with respect to the standard deviation of the predicted parameters as we can see by the lack of visible error bars in \cref{fig:example-nn}.
The solver behaves more inconsistent, generating a tight grouping of parameters for some concentration combinations but failing to do so for others (see \cref{fig:example-solver}).

\subsection{Runtime}
\begin{table}[t]
    \caption{Runtime to calculate parameters for one datapoint measured over three runs. Mean and standard deviation given. Speedup Ours (GPU) vs. L-BFGS-B.}\label{tab:runtime}
    \vspace{2mm}
    \centering
    \begin{tblr}{
        colspec={|l|rr|rr|rr|},
        colsep = 4pt,
        column{3,5,7} = {leftsep=0pt},
        hlines,
        vline{1,2,4,6,8} = {solid},
    }
        & \SetCell[c=2]{c} Lorentzian & & \SetCell[c=2]{c} analytical $Z$ & & \SetCell[c=2]{c} $MTR_{Rex}$ & \\
        Nelder-Mead & 636.90 $\pm$ & 205.27 \si{\ms} & 6374.88 $\pm$ & 420.30 \si{\ms} & 5468.75 $\pm$ & 1470.39 \si{\ms} \\
        Powell & 641.97 $\pm$ & 205.05 \si{\ms} & 2172.11 $\pm$ & 1085.54 \si{\ms} & 7446.31 $\pm$ & 3878.10 \si{\ms} \\
        L-BFGS-B & 149.69 $\pm$ & 35.64 \si{\ms} & 1634.41 $\pm$ & 703.71 \si{\ms} & 4042.41 $\pm$ & 1911.56 \si{\ms} \\
        Ours (GPU) & 9.04 $\pm$ & 23.25 \si{\ms} & 8.64 $\pm$ & 22.14 \si{\ms} & 8.99 $\pm$ & 22.52 \si{\ms} \\
        Ours (CPU) & 92.54 $\pm$ & 16.71 \si{\ms} & 92.99 $\pm$ & 15.22 \si{\ms} & 93.92 $\pm$ & 15.72 \si{\ms} \\
        Speedup & \SetCell[c=2]{r} 16.56 $\times \,\ $ & & \SetCell[c=2]{r} 189.17 $\times \,\ $ & & \SetCell[c=2]{r} 449.66 $\times \,\ $ & \\
    \end{tblr}
\end{table}

We conducted a short runtime evaluation in \cref{tab:runtime}.
Apart from better results, our network is also faster than the solvers even when performing inference on the CPU.
As the number of parameters are in the same range for all three models, the networks inference speed is almost constant between them.
The solvers takes longer as the models get more complex as the model itself needs to be evaluated in each iteration.
Secondly, with increasing model complexity finding a good set of parameters is also harder and, therefore, convergence slower.
For a 3D volume of size $10 \times 10 \times 10$ the speedup would result in computation time of around $4000 \si{\sec}$ for L-BFGS-B vs $9 \si{\sec}$ for our network.

\section{Conclusion}\label{sec:conclusion}
In this work we have presented a new approach to model-based CEST analysis using a transformer-based neural network.
We have demonstrated its capabilities on a phantom dataset and shown that it outperforms a classical iterative L-BFGS-B solver.
It produces more consistent results over different solute concentration combinations.

The encouraging outcomes from phantom datasets serve as a proof of principle, laying the groundwork for the application of this methodology to \textit{in vivo} data. However, transitioning to \textit{in vivo} data presents significant challenges due to the lack of ground truth and the complexity of \textit{in vivo} CEST spectra. Under physiological conditions, the large exchange rates complicate accurate modeling and analysis. Our future work will focus on overcoming these challenges, refining our approach to reliably analyze \textit{in vivo} CEST data, e.g. by developing regularization based on spatial correlations in \textit{in vivo} MRI data.

\section*{Acknowledgements}
This work was supported by the Deutsche Forschungsgemeinschaft (DFG) – CRC 1450 – 431460824.

\bibliographystyle{splncs04}
\bibliography{bibliography}
\end{document}